\documentclass[runningheads]{llncs}

\usepackage{cite}
\usepackage{graphicx}
\usepackage{textcomp}
\usepackage{xcolor}
\usepackage{multirow}
\usepackage{color}
\usepackage{tabularray}
\usepackage{hyperref} 
\usepackage{cleveref}
\usepackage{makecell}
\usepackage{array}
\usepackage{booktabs} 
\usepackage[T1]{fontenc}
\usepackage{float}

\hypersetup{
    colorlinks=true,
    linkcolor=black,
    filecolor=magenta,      
    urlcolor=black,
    citecolor=black,
}
\def\BibTeX{{\rm B\kern-.05em{\sc i\kern-.025em b}\kern-.08em
    T\kern-.1667em\lower.7ex\hbox{E}\kern-.125emX}}

\begin{document}
\title{ForCM: Forest Cover Mapping from Multispectral Sentinel-2 Image by Integrating Deep Learning with Object-Based Image Analysis}
\titlerunning{ForCM}

\author{Maisha Haque\inst{1,*}
\and
Israt Jahan Ayshi\inst{1}
\and
Sadaf M. Anis\inst{1}
\and
Nahian Tasnim\inst{1}
\and
Mithila Moontaha\inst{1}
\and
Md. Sabbir Ahmed\inst{1}
\and
Muhammad Iqbal Hossain\inst{1}
\and
Mohammad Zavid Parvez\inst{2,8,*}
\and
Subrata Chakraborty\inst{3,4,5,*}
\and
Biswajeet Pradhan\inst{4,6}
\and
Biswajit Banik\inst{7}}
\authorrunning{M. Haque et al.}
\institute{BRAC University, Bangladesh
\and
School of Computing, Mathematics and Engineering, Charles Sturt University, Australia
\and
School of Science and Technology, University of New England, Armidale, NSW 2351, Australia
\and
Centre for Advanced Modelling and Geospatial Information Systems (CAMGIS), School of Civil and Environmental Engineering, Faculty of Engineering \& IT, University of Technology Sydney, Sydney, NSW 2007, Australia
\and
Griffith Business School, Griffith University, Nathan, QLD 4111, Australia
\and
Earth Observation Centre, Institute of Climate Change, Universiti Kebangsaan Malaysia (UKM), Bangi 43600, Selangor, Malaysia
\and
Institute of Health and Wellbeing, Federation University Australia, Victoria, Australia\and
School of Accounting, Information Systems and Supply Chain, RMIT University, Australia
}
\maketitle              

\begin{abstract}
This research proposes "ForCM," a novel approach to forest cover mapping that combines Object-Based Image Analysis (OBIA) with Deep Learning (DL) using multispectral Sentinel-2 imagery. The study initially explores the application of several DL models such as UNet, UNet++, ResUNet, AttentionUNet, and ResNet50-Segnet—on high-resolution Sentinel-2 Level 2A satellite images of the Amazon Rainforest. The datasets comprise three primary collections: two sets of 3-band imagery and one set of 4-band imagery. After evaluating the DL models, the most effective ones are individually integrated with the OBIA technique to enhance mapping accuracy. The originality of this work lies in the evaluation of different deep learning models combined with OBIA, and their comparison with traditional OBIA methods. The findings indicate that the proposed "ForCM" method significantly improves forest cover mapping, achieving overall accuracies of 94.54\% with ResUNet-OBIA and 95.64\% with AttentionUNet-OBIA, compared to 92.91\% with the traditional OBIA approach. Furthermore, this research demonstrates the potential of free and user-friendly tools like QGIS for achieving precise mapping within their limitations, supporting global environmental monitoring and conservation efforts.

\end{abstract}
\section{Introduction}
Forests provide crucial ecological services, including carbon sequestration, air quality improvement, biodiversity protection and climate regulation. However, global forestry ecosystems are consistently being threatened by deforestation caused by non-eco-friendly human activities, urbanization, natural calamities, climate change, and wildfires. Sustainable forest management must be administered to safeguard human survival and environmental stability. A key aspect of forest administration strategies is the accurate mapping and monitoring of forest cover, which is critical for mitigating the negative effects of deforestation.

Sentinel-2  satellite imagery is widely used in fields of forest cover mapping including tropical forests \cite{cechim2023object}, mangrove cover \cite{tariq2023modelling}, as well as land cover classification \cite{Zaabar2021-jk} for their multispectral bands. Current approaches to mapping forest cover areas are mostly pixel-based and object-based. Object-Based Image Analysis (OBIA) is well known for its ability to improve segmentation by grouping pixels into meaningful objects compared to pixel-based approaches. Conversely, pixel-based approaches mostly involve machine learning and deep learning architectures, which are well-known for their ability to recognize pixel-level complex patterns and extract relevant features but may struggle with precise object edges.

Even with their advancements, current forest cover mapping approaches face some challenges. They can be costly and inconsistent and the results may vary depending on the analyst’s expertise. Additionally, present methods often face difficulty in reliably detecting and classifying subject edges. Concerning our research scope of forest mapping, such issues can particularly commence in complex landscapes with high occlusion and overlapping canopies. Previous studies have employed OBIA \cite{wahyuni2017forest} and machine learning (ML) \cite{cechim2023object, cissell2021mapping} or DL techniques for example- using  ML with manual refinement \cite{tariq2023modelling} or using deep learning techniques \cite{md2024deforestation, javed2023deep}. Exploring such studies, their findings and limitations, it can be deduced that the effectiveness of OBIA is heavily dependent on unambiguous superior-quality input images and the segmentation procedure, which can lead to overpredicting or underpredicting occasionally in certain scenarios, may it be related to ambiguous input images or the complexity of used architectures. Additionally, the application of OBIA typically requires specialized software and tools which might be costly and unavailable in some regions, complicating the application further.  On the contrary, deep learning algorithms which are capable of recognizing intricate pixel-level patterns, may also fall short in the accurate extraction of object boundaries. 

To address the issues, this research proposes an approach called \emph{ForCM} (\emph{\underline{For}est \underline{C}over \underline{M}apping}), a strategy to integrate OBIA with advanced Deep Learning models to enhance the identification and categorization of projected total forest cover maps utilizing free available tools and resources. Such an integration approach of DL with OBIA has previously been applied in several cases, such as- extracting check dam areas \cite{Li2021-hs} and detecting rock glaciers \cite{Robson2020-gz}, although not yet applied for forest cover mapping. However, ForCM stands out by experimenting with several state-of-the-art DL models (UNet, UNet++, ResUNet, AttentionUNet, and ResNet50-SegNet) for integrating with OBIA and applying the proposed method to high-resolution Sentinet-2A forest imagery. Furthermore, the ForCM approach is tested on more than one dataset including 3-band and 4-band images, targeting a significantly enhanced mapping accuracy. 

The novelty of ForCM is that it outperforms current forest cover mapping techniques \cite{cechim2023object, tariq2023modelling, md2024deforestation , javed2023deep}, by adopting a fusion of the OBIA and DL model that can utilize the strength of both techniques. Moreover, this research uses cost-effective and easily accessible software and tools to make ForCM more convenient. While previous efforts have combined convolutional neural network (CNN) with OBIA for land cover classification \cite{Zaabar2021-jk} and urban tree mapping \cite{Timilsina_Aryal_Kirkpatrick_2020}, ForCM evaluates the integration of advanced DL models with OBIA to achieve higher overall accuracy in mapping forest cover.

The remainder of this paper comprises a brief study of related works, the materials and methods of our proposed approach \emph{ForCM}, result analysis, and the corresponding discussion followed by the conclusion.

\section{Related Works}
The process of combining deep learning models with OBIA is comparatively recently developed. We have found some work based on image processing, mapping with classification and segmentation using
different models where they tried to enhance their accuracy and efficiency. 

The authors of \cite{Cao_Zhang_2020}, introduced tree species classification using airborne orthophotos in Guangxi, China, by combining deep learning
models, U-Net, and ResNet to form ResUNet. This addressed
the gradient degradation problem and enhanced multi-scale spatial feature extraction with an overall accuracy of 87.51\%. 

Also, using high-resolution aerial images in  \cite{alsabhan2022automatic}, automated building extraction in urban areas was proposed by
implementing deep learning models with 90\% classification accuracy. U-Net was used for image segmentation while resolving the vanishing gradient problem and ResNet50 was used as an encoder for U-Net. 

The combination of deep learning models with OBIA can be seen in some research works, but based on the work process, different research papers show different results. In  \cite{Ghorbanzadeh2022-sx} landslide detection maps were generated by using the DL model and OBIA separately and together, they used  ResU-Net, rule-based OBIA on the original image, and a combined ResU-Net-OBIA approach. 

Another work \cite{Timilsina_Aryal_Kirkpatrick_2020} focused on tree cover mapping over the years. It used satellite and LiDAR data from Google Earth imaginaries, with an object-based CNN method with multiresolution segmentation, NDVI, and CHM. The study used different tools like “random point creation” from ArcGIS Pro 2.4 for sample pointing. Though the heatmap was produced by the CNN model, it was further refined by using OBIA. The accuracy was fairly good compared to others, at 96\%. Moreover, the use of accessible sources was inspired by the paper \cite{Zaki_Buchori_Sejati_Liu_2022}, where land cover classification was performed on high-resolution Sentinel-2A images using OBIA aiming for coastal spatial planning. The study indicates the rapid land cover changes. Using QGIS and Orfeo ToolBox with ANN in classified land cover maps with test images helped to obtain 94.50\% accuracy for 2015 images in OBIA. 

The previous studies \cite{Cao_Zhang_2020, alsabhan2022automatic, Ghorbanzadeh2022-sx, Zaki_Buchori_Sejati_Liu_2022} demonstrate several uses and implementations of advanced CNN networks and object-based image analysis techniques. To the best of our knowledge, none of the previous works implemented OBIA with DL models explicitly for forest cover mapping from multi-spectral Sentinel-2 images, similar to our proposed ForCM approach.


\section{Materials and Methods}
 This section comprises our research design and proposed \emph{ForCM} approach, the study area, pre-processing, and model-building processes. 

 \begin{figure}[h]
    \centering
    \includegraphics[width=\textwidth,keepaspectratio]{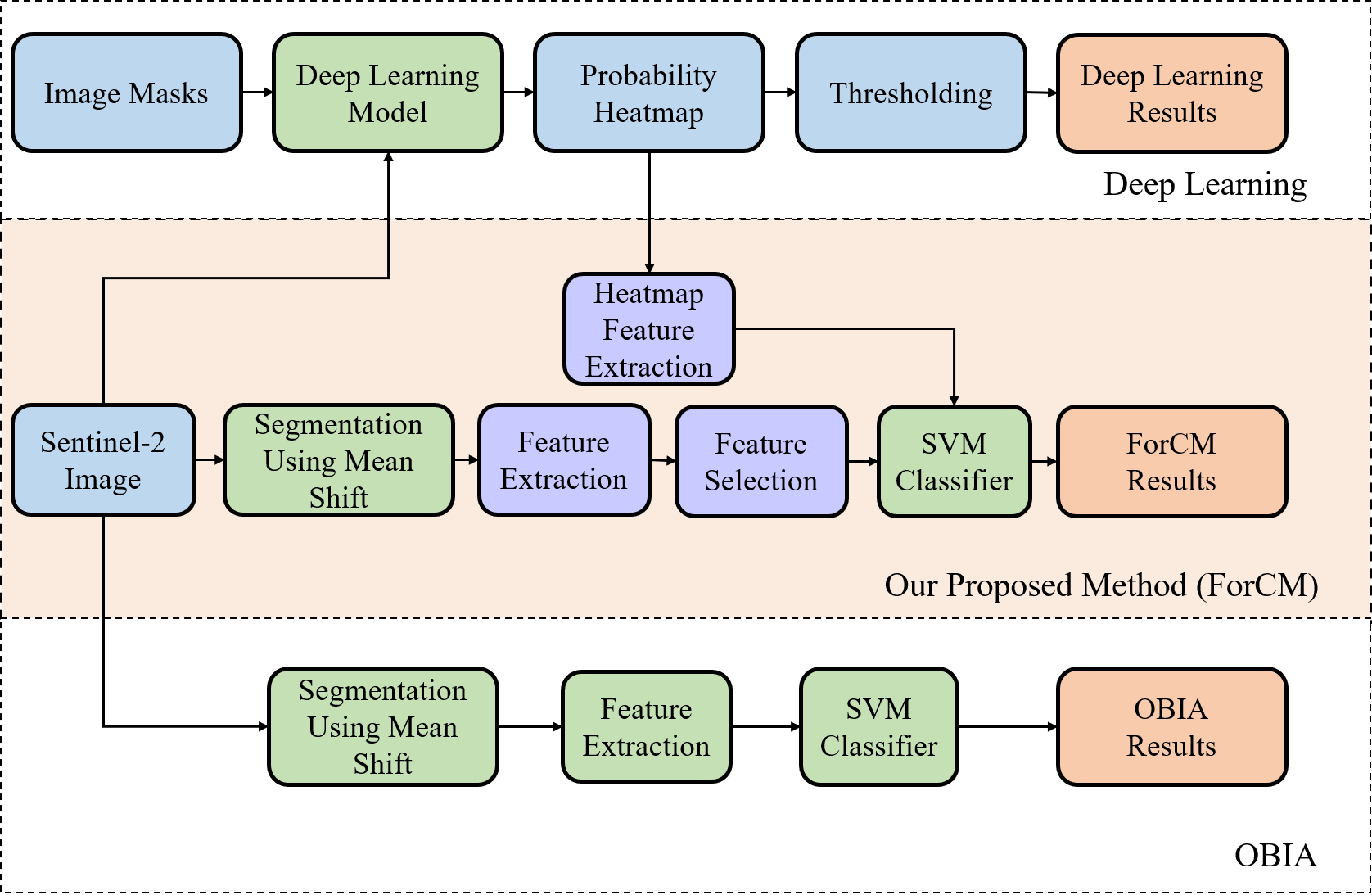}
    \caption{Illustration of the proposed ForCM method of integrating Deep Learning with OBIA for forest Cover Mapping from Multispectral Sentinel-2 Image Datasets.}
    \label{Figure1}
\end{figure}

\subsection{Research Design}
In our research, an attempt is made to analyse the performance of Deep Learning-OBIA for forest cover mapping. An overview of the work of this paper is represented in Figure \ref{Figure1}. First, deep learning models were trained, validated and processed to generate prediction heat maps using the images and ground truth masks of the dataset. On the other hand, the image is segmented in the GIS software. Lastly, a classification algorithm was applied using the image's
spectral features from the predicted heatmap and segmentation output in order to classify the forest cover.

\subsection{Deep Learning Frameworks and GIS software}

  For our model experimentation, the Python 3.9.12 version was used along with Keras of the TensorFlow 2.10.0 framework utilizing "NVIDIA GeForce GTX TITAN X" 12 GB GPU and 32 GB of RAM. Also for software implication an open source and noncommercial, Geographic Information System (GIS) software "QGIS (Quantum GIS)"  version 3.34.5 "Maidenhead" and its plugin "Orfeo Toolbox Provider (OTB)" version 8.1.2 were used to implement the OBIA method.


\subsection{Study Area, Dataset and Preprocessing }

\begin{figure*}[h]
    \centering
    \includegraphics[width=11cm,keepaspectratio]{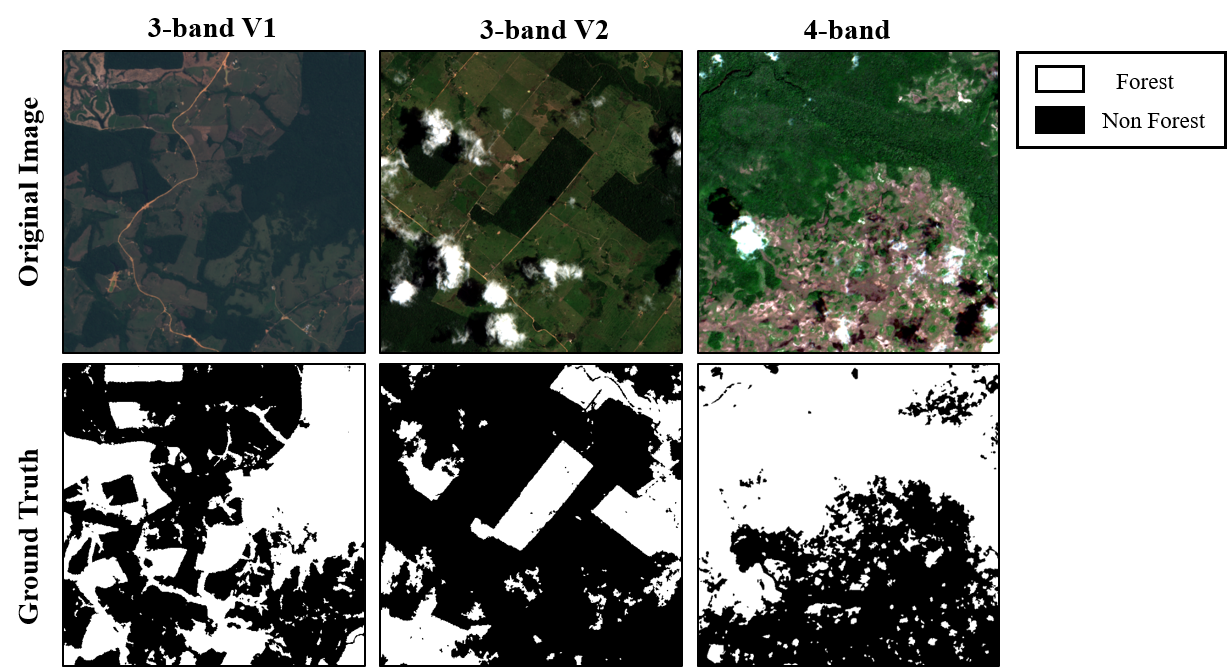}
    \caption{Images from all 3 datasets with their respective ground truth mask.}
    \label{Figure2}
\end{figure*}

This research uses images from the Amazon Rainforest produced by satellite. Three datasets have been used respectively for classifying forest and non-forest, which contain images from Sentinel-2 Level-2A with a spatial resolution of 10 m for four spectral bands (Red, Blue, Green and NIR). All three datasets are open source (satellite imagery) provided by Zenodo organisation \cite{Bragagnolo_Da_Silva_Grzybowski_2019_dataset, bragagnolo2020amazon, Bragagnolo_da_Silva_Grzybowski_2021_dataset}.   In Table \ref{table1} the number of images within these datasets along with the number of train, validation, and test images can be seen. The first two datasets, referred to as the 3-band dataset V1 and 3-band dataset V2 contain images with RGB band and ground truth masks. The 3-band dataset V3 combines selected images of the 3-band dataset V1 and V2. The last dataset contains images with an RGB band with an additional NIR band mentioned as it is the 4-band dataset. All the images are 512 x 512 GeoTIFF images and the ground truth masks for the 3-band dataset are PNG files whereas for the 4-band it is GeoTiff. Due to computational cost, 250 images from the 3-band dataset V2 were selected for training the deep learning models along with the other datasets. Sample images with their ground truth mask from all the datasets can be visualised in Figure \ref{Figure2}. 

For the 3-band datasets, the training images are normalised by splitting their pixel data by 255, then reformed to (512, 512, 3) and converted to float32. The conditioning mask is changed by removing 1 from its pixel values trimming it to (512, 512) restructuring to (512, 512, 1) and casting to int. Moreover, the images of the 4-band dataset are transformed into a float32 category and to ensure that each coaching mask contains just one channel every mask is altered to (1, 512, 512, 1).

\begin{table}
\caption{Number of images for all four datasets.}\label{table1}
\centering
\begin{tabular}{|>{\arraybackslash}p{3cm}|>{\arraybackslash}p{3cm}|>{\arraybackslash}p{3cm}|>{\arraybackslash}p{2.8cm}|}
\hline
\textbf{Dataset Name} &  
\textbf{Training Images} & 
\textbf{Validation Images} & 
\textbf{Test Images}\\
\hline
3-band V1 & 30 & 15 & 15 \\ 

3-band V2 & 1123 & 100 & 100 \\

3-band V3 & 280   & 115 & 100 \\

4-band &  499   & 100 & 20 \\ 
        
\hline
\end{tabular}
\end{table}

\subsection{Deep Learning Models}

Initially, four deep learning models were implemented which are - UNet, UNet++, ResUNet, Attention UNet and ResNet50-Segnet.  All the models are trained and evaluated on all four datasets to compare their performances in terms of accuracy, precision, recall, F1 score, complexity, loss graph and computation time. The first and basic backbone model is  UNet \cite{ronneberger2015u}, which is a very popular model used for semantic segmentation that incorporates convolutional neural networks. Again, Unet++ \cite{zhou2018unet++} is an architecture based on the UNet model where the encoder and decoder sub-networks are connected through nested skip pathways. Furthermore,  ResUNet \cite{zhang2018road} is another architecture that uses UNet’s architecture and combines it with ResNets residual connections, which were created for remote sensing visualisation. The Attention UNet \cite{oktay2018attention} model, which consists of an attention function, enhances segmentation accuracy by perpetually pivoting on key sections while ignoring unnecessary ones. Lastly,  The ResNet50-SegNet \cite{badrinarayanan2017segnet} architecture uses the ResNet50 model and integrates with SegNet frameworks to provide an architecture appropriate for semantic segmentation. 

All the models utilise binary cross entropy as the loss function and sigmoid as the activation function. ResUNet, Attention UNet, and ResNet50-Segnet, from which ResUNet and Attention UNet were chosen to generate heatmaps for further analysis. All the models were run from scratch and tested under identical conditions and a uniform number of epochs for equitable comparison.

 \begin{figure}[h!]
    \centering
    \includegraphics[width=11cm,keepaspectratio]{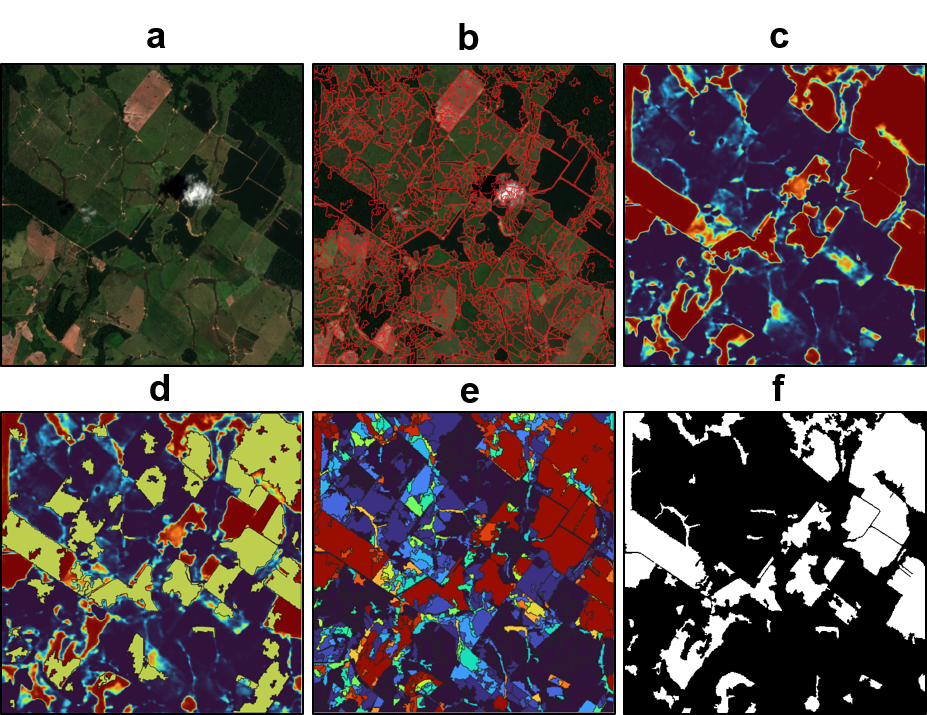}
    \caption{ Input Image (a), Segments after applying Meanshift (b), Generated heatmap from deep leanring Model (ResUNet) where intensity of red shows the probability of forest and intensity of blue shows the opposite (c), randomly selected training segments (in green) consisting features extracted from image objects and heatmap (d), classified image segments with heatmap weights (e), and   binary classified image after thresholding (f)
    }
    \label{Figure3}
\end{figure}

\subsection{OBIA method implementation}
 The OBIA approach can be divided into two important parts: segmentation and classification. For segmenting the image,  an unsupervised segmentation algorithm “mean shift” is used, which breaks the image into objects by clustering the pixels \cite{kotaridis2020object}. The parameters for this algorithm are implemented based on trial and error \cite{rudrapal2015land}.  The parameters used in our work are- spatial radius 5 and range radius 5 and the minimum segment size was chosen based on trial and error. For classification, we used the support vector machine (SVM), which has provided satisfactory results in previous works \cite{hahn2007application, azeez2022integration}. The SVM model in our study was applied with a linear kernel and regularisation parameter 1 to generate the binary classified output.

\subsection{Proposed Approach: ForCM (DL-OBIA)}
Our proposed approach, ForCM, implements a fusion technique by integrating the features of the decisions from the deep learning model with OBIA \cite{picoli2022unsupervised}. ResUNet and Attention UNet models are used for producing the prediction map and like the classic OBIA approach, mean shift and SVM are used for segmentation and classification. Furthermore, free-of-cost software 'QGIS' is used to perform the decision fusion technique to perform DL-OBIA with the OTB plugin. Initially, after segmenting the image into objects, some random training objects are selected. The heatmap weights extracted from the prediction map generated from the deep learning model are combined with the spectral features extracted from the image objects to train the SVM model. The model classifies the objects into different weight values, and these weight values are thresholded into a binary classified output for forest and non-forest cover. In Figure \ref{Figure3}, a sample image and its segments and classified results can be seen. The parameters selected for mean shift and SVM are the same as those selected for the implementation of the OBIA method.

\section{Results and Discussion}
This section covers the result analysis according to quantitative evaluation and comparison figures followed by further discussion.

\subsection{Results}

\begin{figure}[h!]
    \centering
    \includegraphics[width=10cm,keepaspectratio]{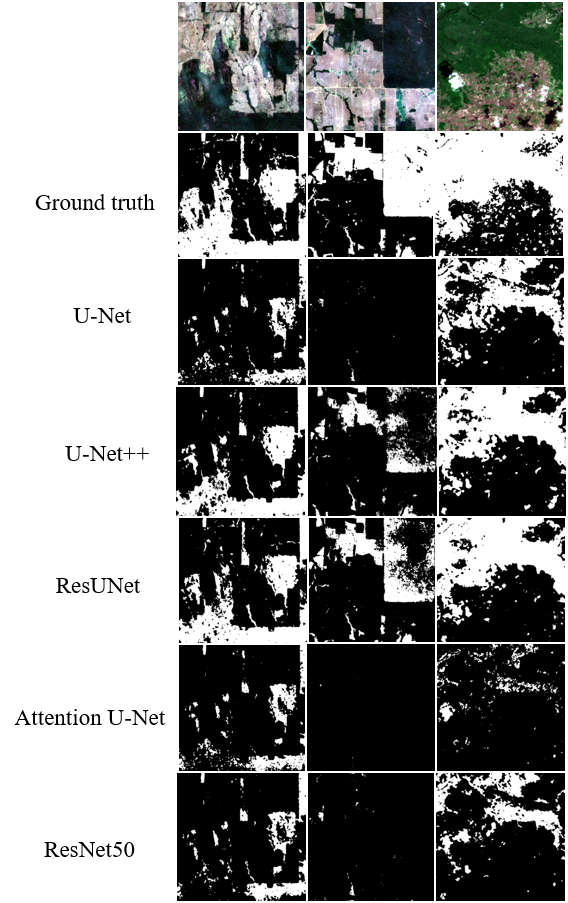}
    \caption{Sample thresholded output mask comparison of the DL models for 4-band dataset.
}
    \label{Figure4}
\end{figure}

\begin{figure}[h!]
    \centering
    \includegraphics[width=9cm,keepaspectratio]{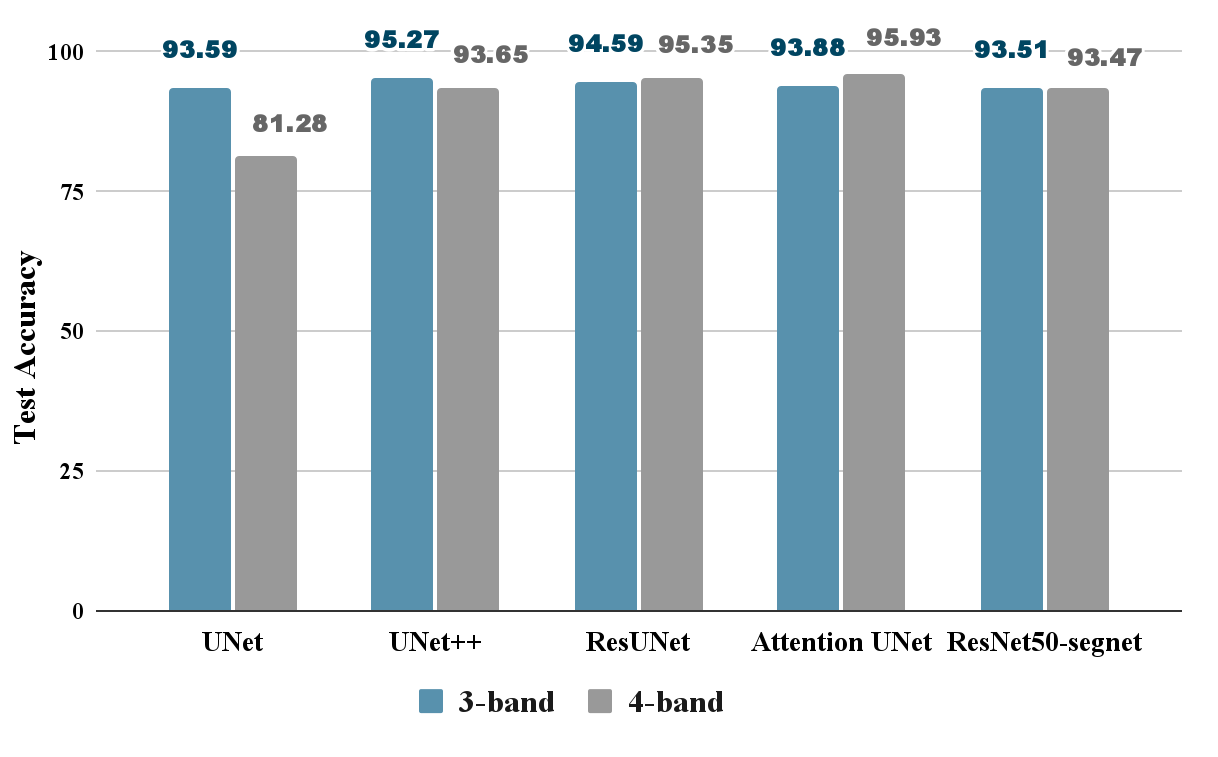}
    \caption{Test Accuracy of the Deep Learning Models of 3-band dataset V3 \&
V4 band dataset.}
    \label{Figure5}
\end{figure}

\begin{figure}[h!]
    \centering
    \includegraphics[width=9cm,keepaspectratio]{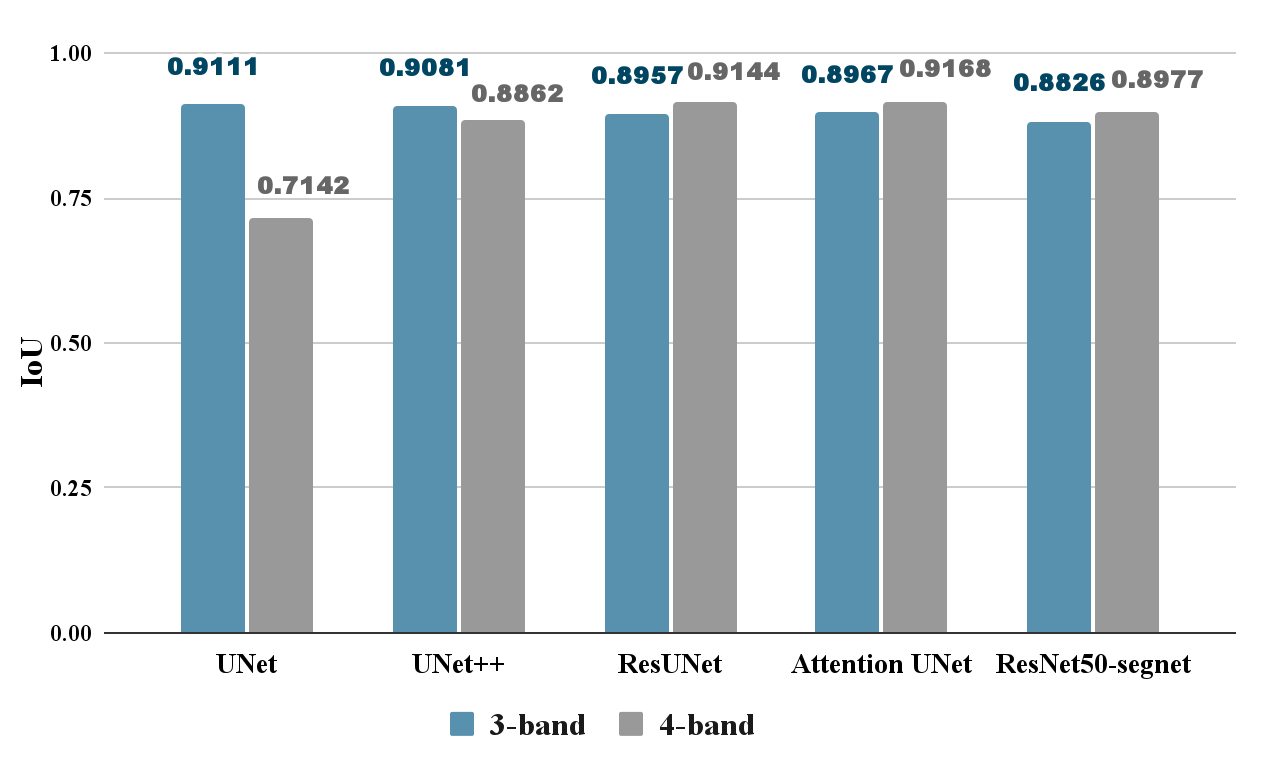}
    \caption{IoU of the Deep Learning Models of 3-band dataset V3 \&
V4 band dataset.}
    \label{Figure6}
\end{figure}

As there are four different datasets considered in this paper, different deep learning model performance measures on all the datasets in terms of accuracy, precision, recall, F1 Score, and test accuracy are recorded in Table \ref{table2}. Additionally, with Table \ref{table3} recording the metric values, the overall results of the proposed combined ForCM (DL-OBIA) approach are further discussed in this section.  
Moreover, Figure \ref{Figure4} includes the comparison of thresholded outputs for all DL models, generated from some example images of 4-band datasets. Besides, bar charts for the model's test accuracy (Figure \ref{Figure5})  and IoU (Figure \ref{Figure6}) can be seen in this section. Lastly, the overall analysis metric values after implementing the OBIA method in the continuation of deep learning are also discussed with visual representations. 

\begin{table*}[htbp]
\caption{Deep Learning Models for Quantitative Evaluation where the best results for each dataset are in bold.
}
\centering

\renewcommand{\arraystretch}{1.5} 
\begin{tabular}{|>{\arraybackslash}p{1.4cm}|>{\arraybackslash}p{1.6cm}|>{\arraybackslash}p{1cm}|>{\arraybackslash}p{1.2cm}|>{\arraybackslash}p{1.2cm}|>{\arraybackslash}p{1.2cm}|>{\arraybackslash}p{1.2cm}|>{\arraybackslash}p{1.2cm}|>{\arraybackslash}p{1.2cm}|}
\hline
\textbf{Model} & \textbf{Amazon Dataset} & \textbf{Epoch} & \textbf{IoU} & \textbf{\makecell{\\Accur \\-acy}} & \textbf{\makecell{\\Preci \\-sion}} & \textbf{Recall} & \textbf{F1 Score} & \textbf{Test Accuracy} \\ 

\hline
\specialrule{.1em}{.3cm}{0em}  

\multirow{4}{*}{UNet} & 3-band V1 & 20 & 0.8948 & 94.15\% & 95.19\% & 94.15\% & 0.9467 & N/A \\ \cline{2-9} 
& 3-band V2 & 10 & 0.9055 & 94.34\% & 95.19\% & 94.34\% & 0.9411 & 93.40\% \\ \cline{2-9} 
& 3-band V3 & 10 & \textbf{0.9111} & 94.00\% & 94.19\% & 94.00\% & 0.9439 & 93.59\% \\ \cline{2-9} 
& 4-band & 10  & 0.7142 &  81.24\% & 82.00\%  & 81.26\%  & 0.8189  & 81.28\%  \\ 
\hline
\specialrule{.1em}{.3cm}{0em}
\multirow{4}{*}{UNet++} & 3-band V1 & 20 & 0.8956 & 94.25\% & 95.44\% & 94.25\% & 0.9491 & N/A \\ \cline{2-9} 
& 3-band V2 & 10 & \textbf{0.9074} & 94.34\% & 95.19\% & 94.34\% & 0.9411 & 93.64\% \\ \cline{2-9} 
& 3-band V3 & 10 & 0.9081 & \textbf{94.66\%} & 95.56\% & \textbf{94.66\%} & 0.9439 & 93.65\% \\ \cline{2-9} 
& 4-band & 10  & 0.8862 & 93.73\% & 94.75\%  & 93.73\%  & 0.9424  & 93.55\%  \\ 
\hline
\specialrule{.1em}{.3cm}{0em}
\multirow{4}{*}{ResUNet} & 3-band V1 & 20 & \textbf{0.9044} & \textbf{94.77\%} & \textbf{95.54\%} & \textbf{94.77\%} & \textbf{0.9515} & N/A \\ \cline{2-9} 
& 3-band V2 & 10 & 0.8998 & \textbf{94.37\%} & 95.13\% & \textbf{94.37\%} & 0.9475 & 94.38\% \\ \cline{2-9} 
& 3-band V3 & 10 & 0.8957 & 94.51\% & \textbf{95.61\%} & 94.51\% & \textbf{0.9456} & \textbf{94.59\%} \\ \cline{2-9} 
& 4-band & 10 & 0.8957 & 93.79\% & 95.37\% & 93.79\% & 0.9557 & 92.34\% \\ 
\hline
\specialrule{.1em}{.3cm}{0em}
\multirow{4}{*}{\makecell{Attention \\ UNet}} & 3-band V1 & 20 & 0.8969 & 93.56\% & 94.32\% & 93.56\% & 0.9510 & N/A \\ \cline{2-9} 
& 3-band V2 & 10 & 0.9001 & 94.28\% & \textbf{95.32\%} & 94.28\% & \textbf{0.9479} & \textbf{94.66\%} \\ \cline{2-9} 
& 3-band V3 & 10 & 0.8967 & 93.41\% & 94.71\% & 93.41\% & 0.9406 & 93.88\% \\ \cline{2-9} 
& 4-band & 10 &\textbf{0.9168} & \textbf{95.47\% }& \textbf{96.14\%} & \textbf{95.47\%} & \textbf{0.9581} &\textbf{95.93\%} \\ 
\hline
\specialrule{.1em}{.3cm}{0em}
\multirow{4}{*}{\makecell{ResNet50 \\-SegNet}} & 3-band V1 & 20 & 0.8977 & 94.25\% & \
95.32\% & 94.25\% & 0.9478 & N/A \\ \cline{2-9} 
& 3-band V2 & 10  & 0.8827 & 93.20\% &94.24\%  & 93.20\% & 0.9372 & 93.47\% \\ \cline{2-9} 
& 3-band V3 & 10 & 0.8826 & 93.24\% & 94.28\% & 93.24\% & 0.9376 & 93.51\% \\ \cline{2-9} 
& 4-band & 10 & 0.8977 & 94.27\% & 95.18\% & 94.39\% & 0.9478 & 93.47\% \\ 
\hline
\end{tabular}

\label{table2}
\end{table*}

Considering various factors like accuracy metrics, training, and validation graph accuracy, prediction mask comparisons with original images and ground truth mask, computational and time complexity, and literature review, two of the above-mentioned DL models (ResUnet and AttentionUnet) heatmap predictions are considered for further Object-based Image Analysis in the next section.

\begin{figure*}[h!]
    \centering
    \includegraphics[width=\textwidth,keepaspectratio]{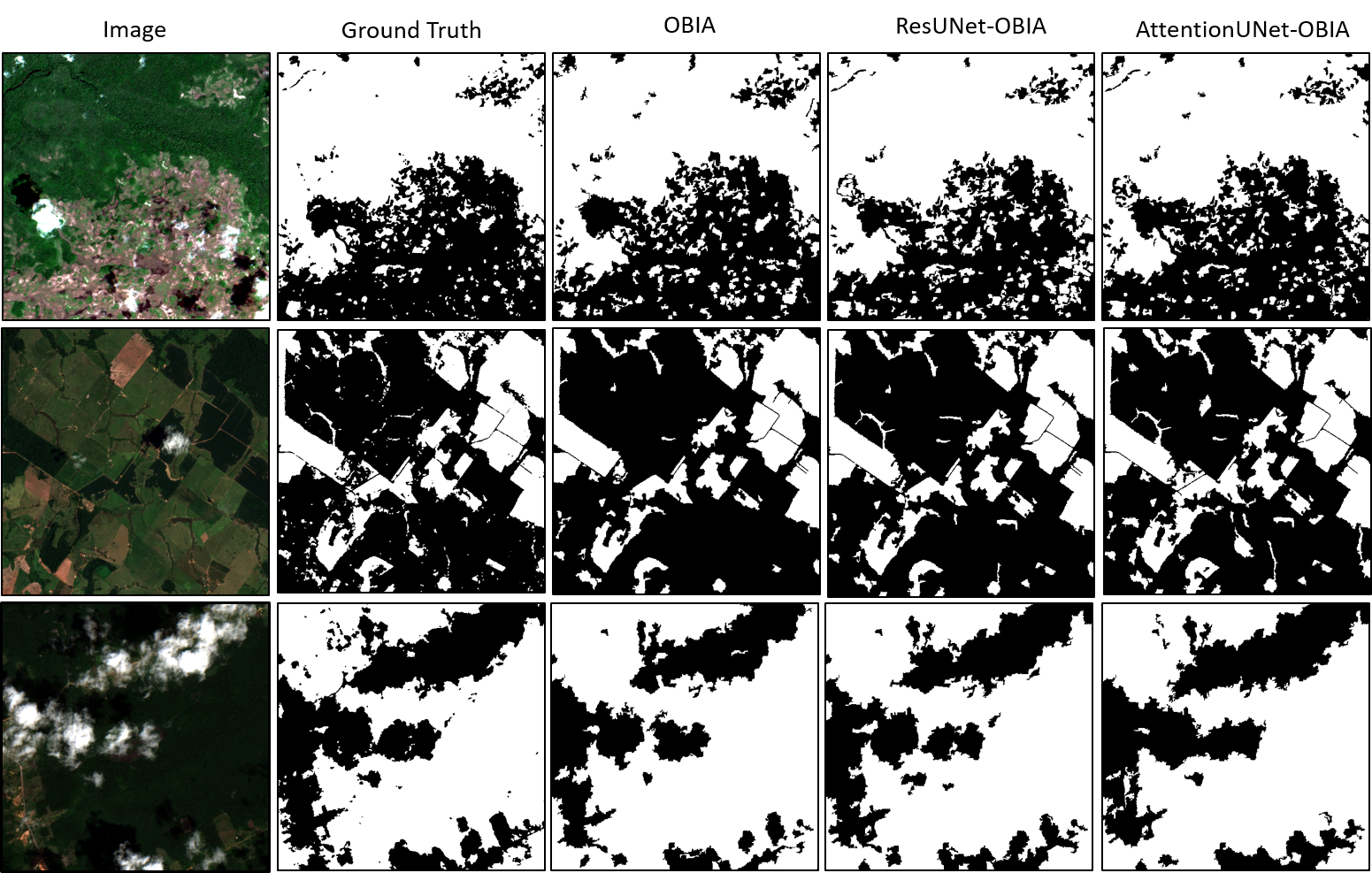}
    \caption{Comparison among classification results of OBIA, ResUNet-OBIA, AttentionUNet-OBIA}
    \label{Figure7}
\end{figure*}

To compare the results of ForCM (DL-OBIA) with traditional OBIA, some randomly selected images and their prediction maps were processed. Considering various factors like accuracy metrics, training, and validation graph accuracy, computational and time complexity, and robustness of the models analysed from previous works ResUNet and Attention UNet were selected for further analysis. The Attention UNet model provides the best result for 4 band model and UNet++ and ResUnet models yield better results for both 3-band images than the rest, which is why prediction map from ResUNet and AttentionUNet models have been utilised to perform our ForCM method. In Figure \ref{Figure7}, the side-by-side visual comparisons between classification results
of OBIA, ResUNet-OBIA, Attention UNet-OBIA and ground truth image can be seen where it can be observed that ForCM has an improved the classified results than traditional OBIA. The classified images of OBIA, ResUNEt-OBIA and AttentionUNet-OBIA have been compared with the ground truth mask using the following metrics-  IoU, overall accuracy, precision, recall and F1 score. In Table \ref{table3}, it can be established that- compared to the effectiveness of basic OBIA, our proposed integrated method has noticeably enhanced outcomes as the average classification accuracy of ResUNet-OBIA yields the most efficient results in terms of IoU  (0.9101), precision (0.9369), recall  (0.9698) and F1 score  (0.9525).

\begin{table}[h]
\caption{Performance evaluation in terms of mean IoU, Overall Accuracy (OA), Precision, Recall and F1 score(F1) (highest in bold) for some randomly selected images from different datasets altogether.}
\renewcommand{\arraystretch}{1.3}
\begin{tabular}{|>{\arraybackslash}p{3.5cm}|>{\arraybackslash}p{1.6cm}|>{\arraybackslash}p{1.6cm}|>{\arraybackslash}p{1.6cm}|>{\arraybackslash}p{1.6cm}|>{\arraybackslash}p{1.6cm}|}
\hline
\textbf{Models} & \textbf{IoU} & \textbf{OA} & \textbf{Precision} & \textbf{Recall} & \textbf{F1} \\ 
\hline

\textbf{OBIA} & {0.8992} & {92.91\%} & {92.43\%} & {95.12\%} & {0.9365} \\ 
\hline
\textbf{ResUNet-OBIA} & \textbf{0.9101} & {94.54\%} & \textbf{93.69\%} & \textbf{96.98\%} & \textbf{0.9525} \\ 
\hline
\textbf{Attention-OBIA} & {0.9064} & \textbf{95.64\%} & {93.32\%} & {96.84\%} & {0.9504} \\ 
\hline

\end{tabular}

\label{table3}
\end{table}

Meanwhile, the integrated AttentionUnet-OBIA achieves the best overall accuracy (0.9564) compared to the rest. It can be observed that the deep learning models enhance the overall outcome of basic OBIA classification as Attention-OBIA increased the IoU of basic OBIA from 0.8992 to 0.9064 while ResUNet-OBIA improved the Overall Accuracy of basic OBIA from 0.9291 to 0.9454. Thus, from overall performance, it can be determined that our proposed method ForCM, especially ResUNet-OBIA significantly outperforms basic OBIA classification.

\subsection{Discussion}

Deep learning strategies have gained popularity in various studies, but merging them with OBIA can improve chronologically comprehensive visualisation and identification using higher-resolution images and projections. Tuning DL architectural factors, such as sample patch measurements and different layer configurations can enhance classification integrity. OBIA technique benefits from improved segmentation parameter enhancement, particularly in segmentation tasks. 

Our process ForCM method involves training four image datasets with four DL models, with ResUNet and AttentionUNet outperforming others. 
Afterward, we trained our OBIA model on the extracted features from the heatmaps produced by these two models. 'Mean Shift' and ‘SVM’ from available algorithms in QGIS, were used for segmentation and classification on selected images respectively, leading to a satisfactory result. Though the ForCM approach has proven to be effective, there are several limitations to be acknowledged. One of the main issues is parameter sensitivity, where the performance can vary widely with different settings of parameters, as the OBIA process needs to be implemented using GIS software, thus needing optimization. Furthermore, the approach also relies on the image quality since low-quality and unclear images may result in less precise segmentation results. On the other hand, while ForCM managed to improve the accuracy using QGIS, better segmentation methods for example multi-resolution segmentation which is offered in paid GIS softwares with more features and likely improved performance in segmentation


\section{Conclusion}

  This research presents "ForCM" as an innovative approach that integrates Object-Based Image Analysis (OBIA) with Deep learning models to improve the accuracy of forest cover mapping using high-resolution Sentinel-2 imagery. ForCM advances the potential of more approachable and precise mapping by addressing the drawbacks of current methods, such as over-prediction, edge detection challenges or inaccessible commercial software constraints. The suggested integration method of DL models such as ResUNet, and AttentionUNet with OBIA has achieved improved accuracy values(94.54\%, 95.64\%), thus contributing substantially to forest monitoring endeavors. Forest cover mapping applications are projected to increase significantly with the rising availability of high-resolution image data sources. Looking ahead, our proposed ForCM strategy can be used to monitor transformations in forest cover over consecutive years, offering a way to calculate the pace of deforestation and environmental transformations. Moreover, future research experimentation of similar detection methodologies can be done using more complex, larger, diverse and higher-quality image datasets that can refine the applicability of our proposed strategy.  In conclusion, the proposed ForCM approach is a powerful procedure for mapping forest cover mapping, contributing to the possibilities of future geospatial monitoring and worldwide sustainable forest management.

\bibliographystyle{splncs04}
\bibliography{reference}

@ARTICLE{Ghorbanzadeh2022-sx,
  title    = "Landslide detection using deep learning and object-based image
              analysis",
  author   = "Ghorbanzadeh, Omid and Shahabi, Hejar and Crivellari, Alessandro
              and Homayouni, Saeid and Blaschke, Thomas and Ghamisi, Pedram",
  journal  = "Landslides",
  volume   =  19,
  number   =  4,
  pages    = "929--939",
  year     =  2022,
  language = "en"
}

@ARTICLE{Li2021-hs,
  title    = "Extracting check dam areas from high‐resolution imagery based on
              the integration of object‐based image analysis and deep learning",
  author   = "Li, Sijin and Xiong, Liyang and Hu, Guanghui and Dang, Weiqin and
              Tang, Guoan and Strobl, Josef",
  journal  = "Land Degrad. Dev.",
  volume   =  32,
  number   =  7,
  pages    = "2303--2317",
  year     =  2021,
  language = "en"
}

@ARTICLE{Robson2020-gz,
  title    = "Automated detection of rock glaciers using deep learning and
              object-based image analysis",
  author   = "Robson, Benjamin Aubrey and Bolch, Tobias and MacDonell, Shelley
              and H{\"o}lbling, Daniel and Rastner, Philipp and Schaffer,
              Nicole",
  journal  = "Remote Sens. Environ.",
  volume   =  250,
  number   =  112033,
  pages    = "112033",
  year     =  2020,
  language = "en"
}

@ARTICLE{Zaabar2021-jk,
  title    = "Assessment of combining convolutional neural networks and Object
              Based Image Analysis to land cover classification using Sentinel
              2 satellite imagery (Tenes region, Algeria)",
  author   = "Zaabar, N and Niculescu, S and Mihoubi, M K",
  journal  = "ISPRS - Int. Arch. Photogramm. Remote Sens. Spat. Inf. Sci.",
  volume   = "XLIII-B3-2021",
  pages    = "383--389",
  year     =  2021,
  language = "en"
}

@article{Zaki_Buchori_Sejati_Liu_2022, title={An object-based image analysis in QGIS for Image Classification and assessment of Coastal Spatial Planning}, volume={25}, DOI={10.1016/j.ejrs.2022.03.002}, number={2}, journal={The Egyptian Journal of Remote Sensing and Space Science}, author={Zaki, Abdurrahman and Buchori, Imam and Sejati, Anang Wahyu and Liu, Yan}, year={2022}, pages={349–359}}

@article{Cao_Zhang_2020, title={An improved Res-UNET model for tree species classification using airborne High-Resolution images}, volume={12}, url={https://doi.org/10.3390/rs12071128}, DOI={10.3390/rs12071128}, number={7}, journal={Remote Sensing}, publisher={Multidisciplinary Digital Publishing Institute}, author={Cao, Kaili and Zhang, Xiaoli}, year={2020}, month={Apr}, pages={1128} }

@article{Timilsina_Aryal_Kirkpatrick_2020, title={Mapping urban tree cover changes using Object-Based Convolution Neural Network (OB-CNN)}, volume={12}, url={https://doi.org/10.3390/rs12183017}, DOI={10.3390/rs12183017}, number={18}, journal={Remote Sensing}, publisher={Multidisciplinary Digital Publishing Institute}, author={Timilsina, Shirisa and Aryal, Jagannath and Kirkpatrick, J. B.}, year={2020}, month={Sep}, pages={3017} }

@misc{Bragagnolo_da_Silva_Grzybowski_2021_dataset, title={Amazon and Atlantic Forest Image datasets for semantic segmentation}, url={https://doi.org/10.5281/zenodo.4498086}, journal={Zenodo}, author={Bragagnolo, Lucimara and da Silva, Roberto Valmir and Grzybowski, José Mario Vicensi}, year={2021}, month={Feb}}

@article{Bragagnolo_Da_Silva_Grzybowski_2019_dataset, title={Amazon Rainforest dataset for semantic segmentation V2}, url={https://zenodo.org/record/3233081}, DOI={10.5281/zenodo.3233081}, journal={Zenodo.org}, author={Bragagnolo, Lucimara and Da Silva, Roberto Valmir and Grzybowski, J.M.V.}, year={2019}, month={May}, language={en} }

@article{zhang2018road,
  title={Road extraction by deep residual u-net},
  author={Zhang, Zhengxin and Liu, Qingjie and Wang, Yunhong},
  journal={IEEE Geoscience and Remote Sensing Letters},
  volume={15},
  number={5},
  pages={749--753},
  year={2018},
  publisher={IEEE}
}

@inproceedings{ronneberger2015u,
  title={U-net: Convolutional networks for biomedical image segmentation},
  author={Ronneberger, Olaf and Fischer, Philipp and Brox, Thomas},
  booktitle={Medical image computing and computer-assisted intervention--MICCAI 2015: 18th international conference, Munich, Germany, October 5-9, 2015, proceedings, part III 18},
  pages={234--241},
  year={2015},
  organization={Springer}
}

@article{oktay2018attention,
  title={Attention u-net: Learning where to look for the pancreas},
  author={Oktay, Ozan and Schlemper, Jo and Folgoc, Loic Le and Lee, Matthew and Heinrich, Mattias and Misawa, Kazunari and Mori, Kensaku and McDonagh, Steven and Hammerla, Nils Y and Kainz, Bernhard and others},
  journal={arXiv preprint arXiv:1804.03999},
  year={2018}
}

@article{badrinarayanan2017segnet,
  title={Segnet: A deep convolutional encoder-decoder architecture for image segmentation},
  author={Badrinarayanan, Vijay and Kendall, Alex and Cipolla, Roberto},
  journal={IEEE transactions on pattern analysis and machine intelligence},
  volume={39},
  number={12},
  pages={2481--2495},
  year={2017},
  publisher={IEEE}
}

@article{cechim2023object,
  title={Object-Based Image Analysis (OBIA) and Machine Learning (ML) Applied to Tropical Forest Mapping Using Sentinel-2},
  author={Cechim Junior, Clovis and Araki, Hideo and de Campos Macedo, Rodrigo},
  journal={Canadian Journal of Remote Sensing},
  volume={49},
  number={1},
  pages={2259504},
  year={2023},
  publisher={Taylor \& Francis}
}

@article{kotaridis2020object,
  title={Object-based image analysis of different spatial resolution satellite imageries in urban and suburban environment},
  author={Kotaridis, Ioannis and Lazaridou, Maria},
  journal={The International Archives of the Photogrammetry, Remote Sensing and Spatial Information Sciences},
  volume={43},
  pages={105--112},
  year={2020},
  publisher={Copernicus Publications G{\"o}ttingen, Germany}
}

@article{rudrapal2015land,
  title={Land cover classification using support vector machine},
  author={Rudrapal, Dhriti and Subhedar, Mansi},
  journal={Int. J. Eng. Res},
  volume={4},
  number={09},
  pages={584--588},
  year={2015}
}

@article{hahn2007application,
  title={Application of Support Vector Machine for Complex Land Cover Classification using Aster and Landsat Data},
  author={Hahn, CLAUDIA and Wijaya, ARIEF and Gloaguen, RICHARD},
  journal={Proceedings of Gemeinsame Jahrestagung der SGPBF, DGPF und OVG, Pubications of DGPF, Muttenz/Basel, Switzerland},
  pages={149--154},
  year={2007}
}

@article{azeez2022integration,
  title={Integration of Object-Based Image Analysis and Convolutional Neural Network for the Classification of High-Resolution Satellite Image: A Comparative Assessment},
  author={Azeez, Omer Saud and Shafri, Helmi ZM and Alias, Aidi Hizami and Haron, Nuzul AB},
  journal={Applied Sciences},
  volume={12},
  number={21},
  pages={10890},
  year={2022},
  publisher={MDPI}
}

@article{picoli2022unsupervised,
  title={Unsupervised Segmentation Of Smallholder Fields In Mozambique Using PlanetScope Imagery},
  author={Picoli, MCA and Radoux, Julien and Tong, Xioaye and Bey, Adia and Rufin, Philippe and Brandt, Martin and Fensholt, Rasmus and Meyfroidt, Patrick},
  journal={The International Archives of the Photogrammetry, Remote Sensing and Spatial Information Sciences},
  volume={43},
  pages={975--981},
  year={2022},
  publisher={Copernicus GmbH}
}

@article{alsabhan2022automatic,
  title={Automatic building extraction on satellite images using Unet and ResNet50},
  author={Alsabhan, Waleed and Alotaiby, Turky and others},
  journal={Computational Intelligence and Neuroscience},
  volume={2022},
  year={2022},
  publisher={Hindawi}
}

@article{tariq2023modelling,
  title={Modelling, mapping and monitoring of forest cover changes, using support vector machine, kernel logistic regression and naive bayes tree models with optical remote sensing data},
  author={Tariq, Aqil and Jiango, Yan and Li, Qingting and Gao, Jianwei and Lu, Linlin and Soufan, Walid and Almutairi, Khalid F and Habib-ur-Rahman, Muhammad},
  journal={Heliyon},
  volume={9},
  number={2},
  year={2023},
  publisher={Elsevier}
}

@article{cissell2021mapping,
  title={Mapping national mangrove cover for Belize using Google Earth Engine and Sentinel-2 imagery},
  author={Cissell, Jordan R and Canty, Steven WJ and Steinberg, Michael K and Simpson, Lora{\'e} T},
  journal={Applied Sciences},
  volume={11},
  number={9},
  pages={4258},
  year={2021},
  publisher={MDPI}
}

@article{md2024deforestation,
  title={Deforestation detection using deep learning-based semantic segmentation techniques: a systematic review},
  author={Md Jelas, Imran and Zulkifley, Mohd Asyraf and Abdullah, Mardina and Spraggon, Martin},
  journal={Frontiers in Forests and Global Change},
  volume={7},
  pages={1300060},
  year={2024},
  publisher={Frontiers Media SA}
}

@article{javed2023deep,
  title={Deep learning-based detection of urban forest cover change along with overall urban changes using very-high-resolution satellite images},
  author={Javed, Aisha and Kim, Taeheon and Lee, Changhui and Oh, Jaehong and Han, Youkyung},
  journal={Remote Sensing},
  volume={15},
  number={17},
  pages={4285},
  year={2023},
  publisher={MDPI}
}

@inproceedings{zhou2018unet++,
  title={Unet++: A nested u-net architecture for medical image segmentation},
  author={Zhou, Zongwei and Rahman Siddiquee, Md Mahfuzur and Tajbakhsh, Nima and Liang, Jianming},
  booktitle={Deep Learning in Medical Image Analysis and Multimodal Learning for Clinical Decision Support: 4th International Workshop, DLMIA 2018, and 8th International Workshop, ML-CDS 2018, Held in Conjunction with MICCAI 2018, Granada, Spain, September 20, 2018, Proceedings 4},
  pages={3--11},
  year={2018},
  organization={Springer}
}

@misc{bragagnolo2020amazon,
  author       = {Bragagnolo, L. and da Silva, R.V. and Grzybowski, J.M.V.},
  title        = {Amazon Rainforest dataset for semantic segmentation V2},
  year         = {2020},
  publisher    = {Zenodo},
  howpublished = {\url{https://zenodo.org/records/3994970}},
}

@inproceedings{wahyuni2017forest,
  title={Forest change analysis using OBIA approach and supervised classification a case study: Kolaka District, South East Sulawesi},
  author={Wahyuni, Rinda},
  booktitle={2017 International conference on advanced computer science and information systems (ICACSIS)},
  pages={105--110},
  year={2017},
  organization={IEEE}
}
\vspace{12pt}

\end{document}